\documentclass[conference]{IEEEtran}
\IEEEoverridecommandlockouts

\usepackage{cite}
\usepackage{amsmath,amssymb,amsfonts}
\usepackage{graphicx}
\usepackage{textcomp}
\usepackage{xcolor}
\def\BibTeX{{\rm B\kern-.05em{\sc i\kern-.025em b}\kern-.08em
    T\kern-.1667em\lower.7ex\hbox{E}\kern-.125emX}}

\usepackage{hyperref}
\usepackage{cleveref}
\usepackage{algorithm}
\usepackage{algorithmicx}
\usepackage{algpseudocode}
\usepackage{float}
\usepackage{multicol}

\begin{document}

\title{An Enhanced Federated Prototype Learning Method under Domain Shift}

\author{\IEEEauthorblockN{Liang Kuang}
\IEEEauthorblockA{\textit{SUSTech}\\
Shenzhen, China \\
12111012@mail.sustech.edu.cn}
\and
\IEEEauthorblockN{Kuangpu Guo}
\IEEEauthorblockA{\textit{USTC}\\
Hefei, China \\
gkp@mail.ustc.edu.cn}
\and
\IEEEauthorblockN{Jian Liang}
\IEEEauthorblockA{\textit{CASIA\&UCAS}\\
Beijing, China \\
liangjian92@gmail.com}
\and
\IEEEauthorblockN{Jianguo Zhang}
\IEEEauthorblockA{\textit{SUSTech}\\
Shenzhen, China \\
zhangjg@sustech.edu.cn}
}

\maketitle

\begin{abstract}
Federated Learning (FL) allows collaborative machine learning training without sharing private data. Numerous studies have shown that one significant factor affecting the performance of federated learning models is the heterogeneity of data across different clients, especially when the data is sampled from various domains. A recent paper introduces variance-aware dual-level prototype clustering and uses a novel $\alpha$-sparsity prototype loss, which increases intra-class similarity and reduces inter-class similarity. To ensure that the features converge within specific clusters, we introduce an improved algorithm, Federated Prototype Learning with Convergent Clusters, abbreviated as FedPLCC. To increase inter-class distances, we weight each prototype with the size of the cluster it represents. To reduce intra-class distances, considering that prototypes with larger distances might come from different domains, we select only a certain proportion of prototypes for the loss function calculation. Evaluations on the Digit-5, Office-10, and DomainNet datasets show that our method performs better than existing approaches.
\end{abstract}


\section{Introduction}

Federated Learning\cite{fedavg} (FL) is an innovative distributed learning framework that allows clients to collaborate in training a global model using their own local datasets, thus maintaining data privacy. FL has several advantages over traditional distributed learning methods as it reduces communication costs and addresses privacy concerns, leading to widespread adoption across various sectors. However, FL faces challenges, particularly concerning data heterogeneity. In FL, clients gather private data from different sources, resulting in non-independent and identically distributed (non-IID) datasets. These non-IID distributions can cause clients to reach their own local optima, potentially deviating from the global objective. As a result, this deviation may hinder convergence rates and reduce overall model performance\cite{flnoniid}.

In FL applications, different types of heterogeneity issues arise\cite{noniidsurvey}. Initially, works were focused on addressing label skew in non-IID (non-identically distributed) data, where the label distribution varies across different clients' datasets. In this type of heterogeneity, the most significant shift occurs in the final layer of the local model, \textit{i.e.} the classifier\cite{ccvr}. Works by \cite{fedbabu,mendieta2022local,fedfa,creff,ccvr} aim to resolve this issue, resulting in faster and more stable convergence as well as higher accuracy. However, as FL algorithms are applied more broadly, more realistic heterogeneity issues are being considered. Recently, some studies have begun to tackle feature skew, where different clients' datasets have different feature representations for the same labels. In this situation, FL algorithms that only optimize the classification head perform poorly because there are substantial differences between the feature extractors trained by different clients. Therefore, it becomes necessary to introduce features into the loss function and update the feature extractor with back-propagation. Works by \cite{tackling,fpl,fedproto} define the average of features of samples with the same label as prototypes and design loss functions to gradually gather output features closer to corresponding prototypes, ensuring the convergence of the feature extractor. These methods work well under feature skew and even label skew.

\begin{figure}[tb]
  \centering
  \includegraphics[width=\linewidth]{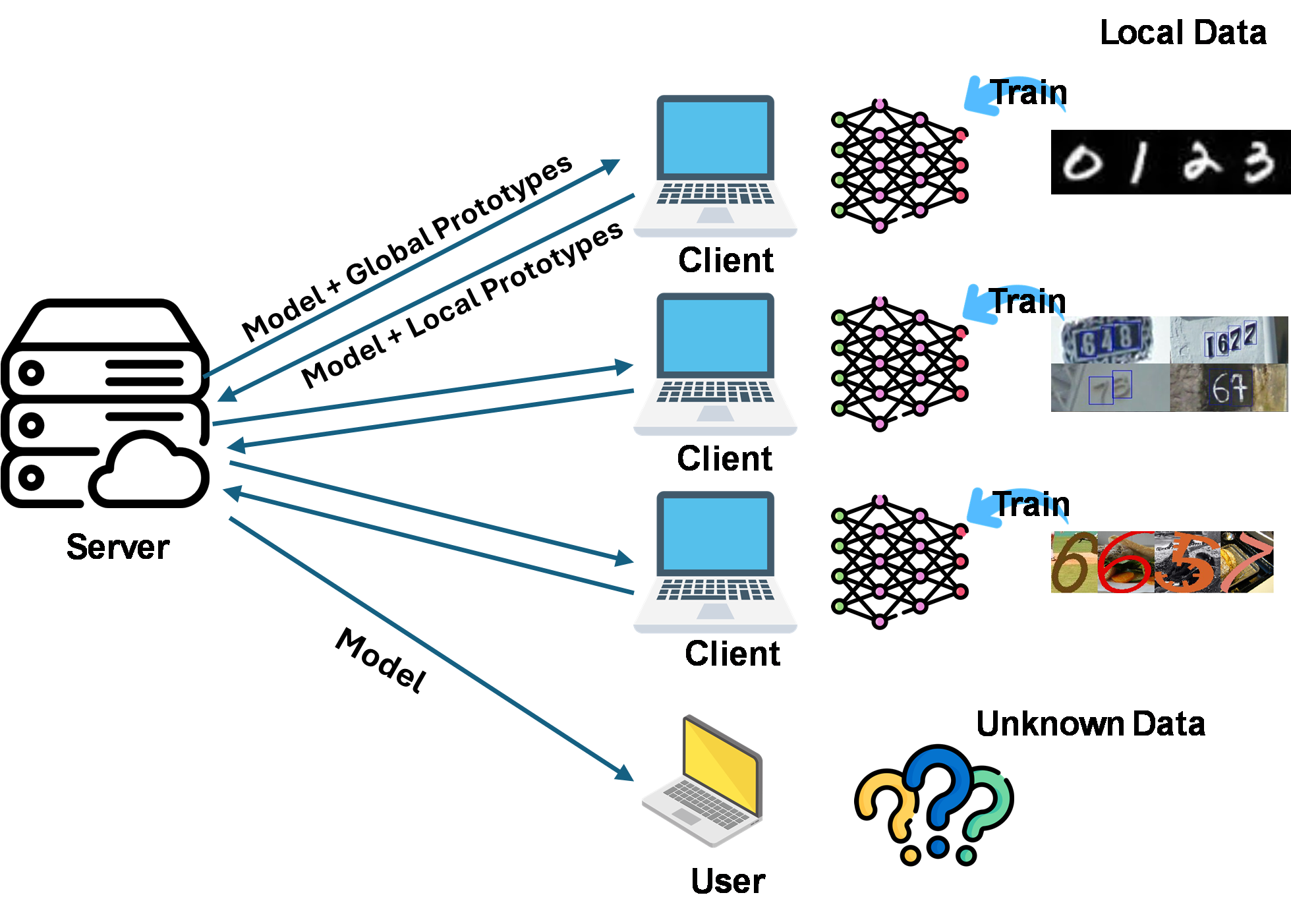}
  \caption{\textbf{Illustration of federated learning with heterogeneous data domains.}}
  \label{fig:1}
\end{figure}

FedPLVM\cite{fedplvm} innovatively redefines the method of computing prototypes by variance-aware dual-level prototype clustering. This approach allows prototypes obtained through clustering to capture richer semantic information. Additionally, FedPLVM employs a novel $\alpha$-sparsity prototype loss, which enhances intra-class similarity and reduces inter-class similarity. Under this framework, FedPLVM narrows the performance gap between easy and hard domains, leading to an overall improvement in average accuracy. However, FedPLVM's dual-level clustering approach has flaws. The local prototypes obtained after the first clustering are treated equally in the second clustering and loss function calculations. The number of samples they represent can vary significantly, contradicting FedPLVM's goal of narrowing the performance gap between the easy and hard domains. Furthermore, when calculating the intra-class loss function, FedPLVM attempts to minimize the distance between each feature and all global prototypes. However, some global prototypes from vastly different domains do not provide a meaningful reference for that feature; thus, using the loss function to minimize these distances can adversely affect the model. 

Considering these factors, we propose an improved algorithm, FedPLCC (short for \textbf{Fed}erated \textbf{P}rototype \textbf{L}earning with \textbf{C}onvergent \textbf{C}lusters). Based on the dual-level clustering framework introduced by FedPLVM, FedPLCC makes two key innovations. \textbf{First}, we consider the number of samples before clustering and incorporate this into the loss function calculation by assigning weights. We notice that a small number of samples with outlier features can form minor outlier clusters during FINCH aggregation, leading to more severe divergence in loss function calculations. \textbf{Second}, when calculating the intra-class loss function, we introduce a hyper-parameter to determine the proportion of prototypes involved in the calculation. Specifically, we select prototypes more similar to the current feature and represent more samples. Within the dual-level clustering framework, some prototypes from other domains may differ significantly from the current feature, not because the model parameters need updating but due to inherent differences in feature expression across domains. Consequently, forcing features to minimize the distance to all prototypes with the same label undermines the model's generalization capability. Consequently, FedPLCC is an algorithm based on Federated Prototype Learning and dual-level clustering. By carefully considering the clustering convergence process, we have endowed it with better generalization capabilities and improved accuracy. Our main contributions are outlined as follows:

\begin{itemize}
    \item This study delves into federated learning with domain shift, summarizing several existing Federated Prototype Learning methods and rethinking their clustering components. We point out that the coarse handling of clustering in existing methods may lead to slower convergence speeds or a decline in final accuracy.
    \item To address this issue, we introduce a new method, FedPLCC. This method builds on the previously established dual-level prototype clustering framework but more accurately captures the local feature distribution by calculating prototype weights. Additionally, in our loss function calculation, we introduce constraints to prevent features from being forced to align with significantly dissimilar prototypes.
    \item Extensive experiments conducted on the Digit-5\cite{digit5}, Office-10\cite{office10}, and DomainNet\cite{domainnet} datasets demonstrate the superior performance of our proposed method when compared with multiple state-of-the-art approaches. All experiment codes will be available on GitHub \textbf{after} the publication of this paper.
\end{itemize}

\section{Related Work}

\subsection{Federated Learning}

Federated Learning (FL) aims to develop a global model through collaboration among multiple clients while protecting their data privacy. FedAvg\cite{fedavg}, the pioneering work in FL, demonstrates that this approach has advantages in terms of privacy and communication efficiency by aggregating local model parameters to train a global model. The challenge of data heterogeneity in FL typically manifests as clients possessing non-IID (independent and identically distributed) data, including label and feature skew. Earlier works recognize that label skew reduces the models' accuracy and attempt to solve this issue. For example, \cite{fedhet,fedhet2} use regularization terms to enhance global model performance. \cite{ccvr,fedbabu,tackling} optimize the classification heads to improve performance. Recent works begin to address feature skew. Some utilize prototypes as global information and clustering them. However, these studies have directly used the FINCH algorithm for clustering without carefully examining the actual impact of the clustering process in FL or exploring how to optimize it. 

\subsection{Prototype Learning}

Prototype learning is a machine learning approach that involves using representative examples of different classes, prototypes, in the learning process. Prototype learning has been extensively explored in various tasks such as transfer learning\cite{pltrans1}, few-shot learning\cite{plfewshot1,plfewshot2}, zero-shot learning\cite{plzeroshot1}, and unsupervised learning\cite{plunsuper1,plunsuper2}. In the FL literature, prototypes are used to abstract knowledge while preserving privacy. For example, FedProto\cite{fedproto} and FedProc\cite{fedproc} align features with global prototypes, CCVR\cite{ccvr} collects Gaussian statistics of clients' data and then uses a Gaussian Mixture Model to generate virtual features. FPL\cite{fpl} clusters local prototypes and calculates unbiased global prototypes to address the issue of clients' data coming from different domains. FedPLVM\cite{fedplvm}, based on FPL, introduces a dual-level clustering framework and $\alpha$-sparsity to reduce the intra-class distances and increase the inter-class distances. Many use clustering algorithms, FINCH\cite{finch} for example, to aggregate local prototypes into global prototypes. Our work examines the clustering process of FL in detail, optimizing the aggregation process and the loss calculation method, thereby improving the model's accuracy.

\subsection{Contrastive Learning}

Contrastive learning has recently emerged as a promising direction in self-supervised learning, achieving competitive results comparable to supervised learning. The primary idea is to bring similar data points closer in the representation space while pushing dissimilar ones apart. A classic work\cite{conlearn} constructs positive and negative pairs for each sample and applies the InfoNCE loss to compare these pairs. Contrastive learning can also be used under fully supervised settings, utilizing both label information and contrastive methods\cite{supervisedcl}. Some works\cite{fpl,fedplvm}, as well as ours, apply contrastive learning to local training of federated learning to enhance performance.

\section{Methodology}

\subsection{Preliminaries}

We follow the classic FL scenario. There are $K$ clients communicating with one server to train an ML model together without sharing their local training data, denoted by $\mathcal{D}_k=\{\mathbf{x_i},y_i\}_i^{N_k}$ for client $k$. The global objective of FL can be formulated as:

\begin{equation}
    \min_w\sum_{k=1}^K\frac{N_k}{N}\mathcal{L}_k(w;\mathcal{D}_k),
\end{equation} where $\mathcal{L}_k$ is the local loss function for client $k$, $w$ denotes the shared global model and $N=\sum_{k=1}^KN_k$ denotes the total number of samples among all clients.

Domain shift exists among clients in heterogeneous federated learning. The conditional feature distribution $P(\mathbf{x}|y)$ varies across clients
while $P(y)$ is consistent, i.e. $P_m(\mathbf{x}|y)\not=P_n(\mathbf{x}|y)\ (P_m(y)=P_n(y))$ for any two clients $m$ and $n$.

\subsection{Federated Prototype Learning}

The work by \cite{ccvr} first points out that the last layer of the model, the classifier, biases the most in heterogeneous federated learning, so they divide the classification network into two parts: the feature extractor and the classifier. The feature extractor $h:\mathbb{R}^V\to\mathbb{R}^D$ maps a sample $\mathbf{x}\in\mathbb{R}^V$ to its feature vector $\mathbf{z}=h(\mathbf{x})\in\mathbb{R}^D$, then the classifier $f:\mathbb{R}^D\to\mathbb{R}^M$ outputs the $M$-class prediction $f(\mathbf{z})=f(h(\mathbf{x}))\in\mathbb{R}^M$. Some previous works \cite{ccvr,fedbabu} adjust the classifier to improve accuracy. In contrast, Federated Prototype Learning methods utilize contrastive methods to optimize the feature extractor. Clients generate local prototypes for each class with the feature vectors of their local samples and share them with the server, and then the server generates global prototypes with the local prototypes collected from the clients. Formally, \begin{equation} \label{equa:loc_clu} \mathcal{P}_k^m=\{p_{k,j}^m\}_{j=1}^{J_k^m}\overset{\text{Algorithm}_l}{\xleftarrow{\hspace{1.1cm}}}\{h(\mathbf{x}_i)|(\mathbf{x}_i,y_i)\in\mathcal{D}^m_k\}, \end{equation} \begin{equation} \label{equa:glo_clu} \mathcal{G}^m=\{g_j^m\}_{j=1}^{C^m}\overset{\text{Algorithm}_g}{\xleftarrow{\hspace{1.1cm}}}\mathcal{P}^m=\{\mathcal{P}^m_k\}^K_{k=1}, \end{equation} where $\mathcal{P}_k^m$ and $\mathcal{G}^m$ represents local prototypes and global prototypes of class $m$ on client $k$ respectively, with size $J_k^m$ and $C^m$. $\text{Algorithm}_l$ and $\text{Algorithm}_g$, representing local clustering algorithm and global clustering algorithm respectively, are defined by the specific FPL algorithm. In the classic work, FedPL\cite{fpl}, $\text{Algorithm}_l$ is averaging and $\text{Algorithm}_g$ is FINCH\cite{finch}, so $J_k^m=1$. In FedPLVM\cite{fedplvm}, both $\text{Algorithm}_l$ and $\text{Algorithm}_g$ are FINCH, which is called dual-level prototype generation. The main purpose is to alleviate the training inequality between easy domains and hard domains. In our study, both $\text{Algorithm}_l$ and $\text{Algorithm}_g$ are $\text{FINCH}^*$, where the prototypes are assigned weights, and the weight of each cluster is the sum of the weights of all its prototypes. We share a similar objective with FedPLVM, but we focus more on the details of the clustering process.

\begin{figure}[tb]
  \centering
  \includegraphics[width=\linewidth]{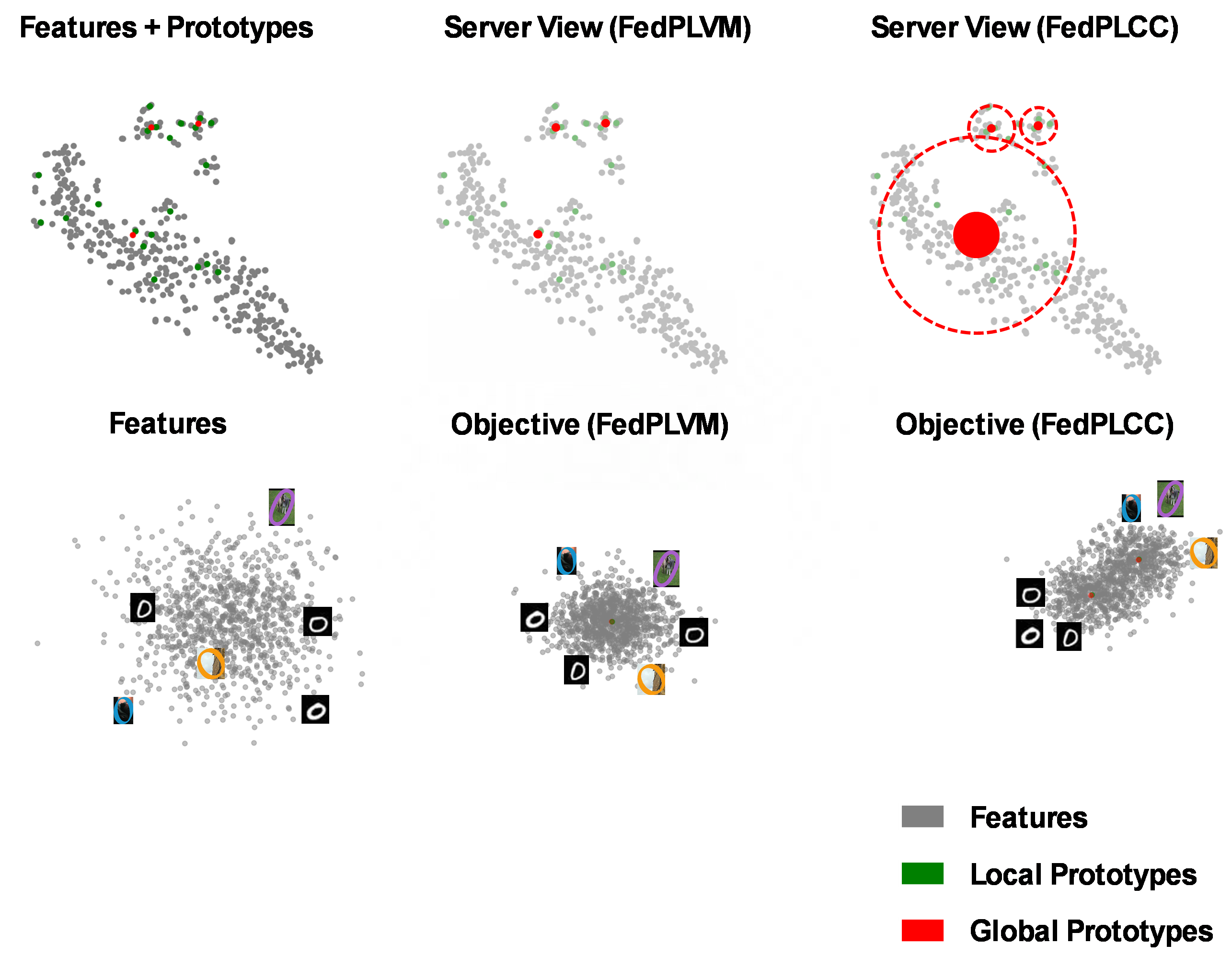}
  \caption{\textbf{Upper row: visualization of features, local prototypes and global prototypes at a specific epoch.} It can be observed that after clustering, there may be significant size differences between different clusters. By assigning weights to local prototypes and global prototypes, we enable the server to better understand the local data distribution. \textbf{Lower row: visualization of the objective of the two methods.} FedPLVM pulls all features inward, whereas FedPLCC attempts to pull features toward several more similar prototypes. While this may increase the overall variance, it can reduce the variance within each cluster.}
  \label{fig:2}
\end{figure}

\subsection{FedPLCC: FedPL with Convergent Clusters}
\label{sec:fedplcc}

We trained a ResNet10\cite{resnet10} model on Digit-5\cite{digit5} with FedPLVM\cite{fedplvm} for several epochs, then used t-SNE to visualize the local features, local prototypes, and global prototypes of label $0$ at a specific epoch, see the upper row in \cref{fig:2}. We have balanced the number of training samples among all clients so that the large differences between clusters do not originate from disparities in the size of local datasets. In fact, a small number of samples might significantly differ from others in the same domain after feature extraction, forming outliers. Due to the parameter-free nature of the FINCH\cite{finch} clustering algorithm, there is a likelihood of forming outlier clusters in such cases. To address this issue, we assign a weight to each prototype. Specifically, whether for local clustering or global clustering, we record the number of samples represented by each prototype when executing the FINCH algorithm and perform a normalization by sum before applying them to loss function calculation, preventing the data volume from affecting the gradient descent step size.

A more important observation here is that we should not expect the model's output features to resemble every prototype. In heterogeneous federated learning, there may be significant differences between domains. However, if features extracted from different domains converge into independent clusters, the model's accuracy can still be improved. Conversely, forcibly bringing features from highly disparate domains into a single cluster may actually reduce the model's accuracy. The difference between the objectives of the two methods is shown in the lower row in \cref{fig:2}. We introduced a top-k mechanism when calculating the intra-class loss function. Specifically, we set a hyper-parameter $\phi\in(0,1]$, which indicates that after sorting the prototypes by their weighted similarity to the current feature, we only take a top proportion ($\phi$) of terms to participate in the loss function calculation. The lower-ranked terms, either not similar to the current feature or represent fewer samples, should not be forcibly aligned with the current feature.

Based on the two ideas above, we redesigned the loss function for FedPLCC:

\begin{equation}
\label{equa:loc_loss}
    \mathcal{L}_{\text{local}}=\mathcal{L}_{\text{CE}}+\lambda_1\mathcal{L}_{\text{contra}}+\lambda_2\mathcal{L}_{\text{corr}},
\end{equation} where $\lambda_1$ and $\lambda_2$ are hyper-parameters balancing the label information loss and contrastive information loss.

The first term is a Cross-Entropy (CE) loss\cite{ce} to train the classifier to yield correct prediction results, which can be formulated as:

\begin{equation}
    \mathcal{L}_{\text{CE}} = \sum_{(\mathbf{x}_i, y_i) \in \mathcal{D}_k} -\mathbf{1}_{y_i} \log(f(h(\mathbf{x}_i))).
\end{equation}

The second term is the contrastive term that pushes prototypes with different labels farther:

\begin{equation}
    \mathcal{L}_{\text{contra}} = -\log \frac{\sum\limits_{g^{y_i} \in \mathcal{G}^{y_i}} \exp(s_{\alpha}(h(\mathbf{x}_i), g^{y_i}) / \tau)\times W_{g^{y_i}}}{\sum\limits_{g \in \mathcal{G}} \exp(s_{\alpha}(h(\mathbf{x}_i), g) / \tau)\times W_{g}},
\end{equation} where \begin{equation} \label{equa:alpha_sparsity}
    s_{\alpha}(h(\mathbf{x}_i), g^m) = \left( \frac{h(\mathbf{x}_i)}{\|h(\mathbf{x}_i)\|} \cdot \frac{g^m}{\|g^m\|} \right)^{\alpha}, 
\end{equation} is called $\alpha$-sparsity\cite{fedplvm}, $\mathcal{G}=\{\mathcal{G}^m\}_{m=1}^M$ is the set of all global clustered prototypes, $\tau$ is the temperature hyper-parameter controlling the strength of the similarity concentration\cite{temp}, and $w_{g^m}$ is the weight of a prototype.

The third term is the contrastive term that gathers the more similar prototypes with the same label:

\begin{equation}
\label{equa:loss_corr}
    \mathcal{L}_{\text{corr}}=-\sum\limits_{g^{y_i} \in \mathcal{G}^{y_i}}^{\text{top k}}(s_{\alpha}(h(\mathbf{x}_i),g^{y_i})\times W_{g^{y_i}}),
\end{equation} where $\overset{\text{top k}}{\sum}$ is a function which takes the hyper-parameter $\phi$ and a set $S$ of $N$ elements as input and returns the sum of the largest $\lceil\phi\times N\rceil$ elements of $S$.

After local training, clients generate local prototypes and corresponding local weights with $\text{FINCH}^*$, and the server aggregates them and generates global prototypes and corresponding global weights:

\begin{equation}
\label{equa:loc_clu_}
    (\mathcal{P}_k^m,\mathcal{W}_{local,k}^m)\overset{\text{FINCH}^*}{\xleftarrow{\hspace{0.65cm}}}(\{h(\mathbf{x}_i)|(\mathbf{x}_i,y_i)\in\mathcal{D}^m_k\},\{1\}_{i=1}^{N_k^m}),
\end{equation}

\begin{equation}
\label{equa:glo_clu_}
    (\mathcal{G}^m,\mathcal{W}_{global}^m)\overset{\text{FINCH}^*}{\xleftarrow{\hspace{0.65cm}}}(\mathcal{P}^m,\mathcal{W}_{local}^m),
\end{equation} where $\mathcal{W}^m_{local,k}=\{W_{p^m_{k,j}}\}_{j=1}^{J_k^m}$ is the local weights of client $k$, $\mathcal{W}_{local}^m=\{\mathcal{W}^m_{local,k}\}_{k=1}^K$ is the aggregated local weights, and $\mathcal{W}_{global}^m=\{W_{g^m_{j}}\}_{j=1}^{C_{m}}$ is the global weights. $\text{FINCH}^*$ takes vectors and their weights as inputs, clusters the vectors with FINCH and sums up the weights of vectors in each cluster as the weight of the cluster. Finally, the server normalizes the weights by sum:

\begin{equation}
\label{equa:nor_glo_w}
    W_{g_j^m}\leftarrow\frac{W_{g_j^m}}{\sum_{j'=1}^{C^m}W_{g_{j'}^m}}.
\end{equation}

The pseudocode for FedPLCC is summarized as \cref{alg:fedplcc}. We have also summarized all the variables and notations used in the algorithm in \cref{var_table}. Only the \textcolor{blue}{blue} items are sent from clients to the server, while only the \textcolor{red}{red} items are sent back from the server to clients. All of them are either weighted clustered prototypes, rather than raw features, or models, which protects the client's privacy.

\renewcommand{\arraystretch}{1.25}
\begin{table}[!tb]
\centering
\caption{\textbf{Variables, notations and corresponding meanings.} Some temporary variables are not mentioned, but they are easy to understand. Details (including the explanation of the colors) in \cref{sec:fedplcc}.}
\label{var_table}
\begin{tabular}{r|p{0.6\linewidth}}
\hline
Variable or Notation & Meaning \\
\hline\multicolumn{2}{l}{\textbf{Problem Setting and Model Definition}}\\ \hline
$K,k$ & The number of clients and one of the clients \\
$M,m$ & The number of classes and one of the classes \\
$\mathcal{D}_k,N_k$ & Client $k$'s training dataset and its size \\
$\color{red}{w},\color{blue}{w_k}$ & The global model and client $k$'s local model \\
$\mathbf{x},y,\mathbf{z}$ & A sample, its label (if available) and its feature \\
$h,f$ & The feature extractor and the classifier, where $w(\mathbf{x})=f(h(\mathbf{x}))=f(\mathbf{z})$ \\
$T,t$ & The number of communication rounds and current round \\
$E,e$ & The number of local training epoch and current epoch \\
\hline\multicolumn{2}{l}{\textbf{Federated Prototype Learning}}\\ \hline
$\color{blue}{\mathcal{P}_k^m,\mathcal{W}^m_{local,k},J_k^m}$ & Sets of local prototypes, corresponding weights and sizes of the sets \\
$\color{red}{\mathcal{G}^m,\mathcal{W}^m_{global},C^m}$ & Sets of global prototypes, corresponding weights and sizes of the sets\\
$\mathcal{L}_{\text{local}},\mathcal{L}_{\text{CE}},\mathcal{L}_{\text{contra}},\mathcal{L}_{\text{corr}}$ & Loss functions \\
$s_\alpha$ & $\alpha$-sparsity \cite{fedplvm} \\
$\alpha,\tau,\lambda_1,\lambda_2,\phi$ & Hyper-parameters used in \cref{equa:alpha_sparsity,equa:loc_loss,equa:loss_corr} \\
\hline
\end{tabular}
\end{table}

\begin{algorithm}[tb]
\caption{FedPLCC}
\label{alg:fedplcc}
\begin{algorithmic}[1]
\State \textbf{Input:} Communication rounds $T$, local training epochs $E$, number of classes $M$, number of clients $K$, private dataset $D_k$
\State \textbf{Output:} Global model $w^{T+1}$

\Statex \textbf{\textcolor{red}{Server Aggregation:}}
\For{$t = 1, 2, \ldots, T$}
    \For{$k = 1, 2, \ldots, K$}
        \State Collect local models, local prototypes, and local prototype weights by
        \Statex $w_{k, E+1}^t, \mathcal{P}_k$, $\mathcal{W}_{local,k} \leftarrow$ \textcolor{blue}{Local Update} $(k, w^t, \mathcal{G}, \mathcal{W}_{global})$
    \EndFor
    \For{$m = 1, 2, \ldots, M$}
        \State Aggregate collected prototypes $\mathcal{P}^m$ and weights $\mathcal{W}^m_{local}$
        \State Generate global prototypes $\mathcal{G}^m$ and weights $\mathcal{W}^m_{global}$ by \cref{equa:glo_clu_}
        \State Normalize global prototypes $\mathcal{W}^m_{global}$ by \cref{equa:nor_glo_w}
    \EndFor
    \State Aggregate global model $w^{t+1} = \sum_{k=1}^K \frac{N_k}{N} w_{k, E+1}^t$
\EndFor
\State Return $w^{T+1}$

\Statex \textbf{\textcolor{blue}{Local Update}}$(k, w^t, \mathcal{G}, \mathcal{W}_{global}):$
\State $w_{k,1}^t \leftarrow w^t$
\For{$e = 1, 2, \ldots, E$}
    \State Update $w_{k,e+1}^t$ from $w_{k,e}^t$ using $\mathcal{G},\mathcal{W}_{global}$ by \cref{equa:loc_loss}
\EndFor
\State Compute local feature vectors $\{h(\mathbf{x}_i)|(\mathbf{x}_i,y_i)\in\mathcal{D}^m_k\}$
\For{$m = 1, 2, \ldots, M$}
    \State Generate local prototypes $\mathcal{P}^m_k$ and weights $\mathcal{W}^m_{local,k}$ by \cref{equa:loc_clu_}
\EndFor
\State Return $w_{k, E+1}^t, \mathcal{P}_k$, $\mathcal{W}_{local,k}$
\end{algorithmic}
\end{algorithm}


\setlength{\tabcolsep}{5pt}

\begin{table*}[!tb]
\centering
\caption{\textbf{Performance comparison on Digit-5.} Avg means the average results among all clients. The best and second-best performances are highlighted via \textbf{bold} and \underline{underline}, respectively. Same for \cref{table2,table3}. Details in \cref{sec:perf_cmp}.}
\label{table1}
\begin{tabular}{l|c|c|c|c|c|c|c}
\hline
Methods & MNIST & USPS & SVHN & Synth & MNIST-M & Avg. & $\Delta$ \\ \hline
FedAvg & 90.10 & 85.05 & 41.66 & 35.20 & 49.96 & 60.394 & - \\
MOON & 79.80 & 83.26 & 29.58 & 25.70 & 40.77 & 51.822 & -8.572 \\
FedProx & 88.47 & 83.56 & 42.47 & 34.65 & 50.36 & 59.902 & -0.492 \\
FedProto & 91.58 & 82.36 & 45.72 & 39.20 & 52.84 & 62.34 & +1.946 \\
FPL & 93.27 & 88.69 & 54.97 & 55.60 & 62.62 & 71.03 & +10.636 \\
FCCL+ & \underline{94.49} & \underline{91.08} & 49.77 & 55.60 & 64.87 & 71.162 & +10.768 \\
FedPLVM & 88.15 & 88.84 & \underline{61.08} & \underline{64.55} & \underline{74.99} & \underline{75.522} & \underline{+15.128} \\
\textbf{Ours} & \textbf{94.92} & \textbf{91.98} & \textbf{66.33} & \textbf{69.65} & \textbf{77.58} & \textbf{80.092} & \textbf{+19.698} \\
\hline
\end{tabular}
\end{table*}

\begin{table*}[!tb]
\centering
\caption{\textbf{Performance comparison on Office-10.} Details in \cref{sec:perf_cmp}.}
\label{table2}
\begin{tabular}{l|c|c|c|c|c|c}
\hline
Methods & Caltech & Amazon & Webcam & DSLR & Avg. & $\Delta$ \\ \hline
FedAvg & 62.50 & 83.16 & 72.41 & 53.33 & 67.85 & - \\
MOON & 51.79 & 72.63 & 77.59 & 36.67 & 59.67 & -8.18 \\
FCCL+ & 57.14 & 77.89 & 75.86 & 53.33 & 66.055 & -1.795 \\
FedProx & 54.46 & 75.79 & \textbf{86.21} & 56.67 & 68.282 & +0.432 \\
FedProto & 57.59 & 76.32 & \underline{84.48} & 60.00 & 69.598 & +1.748 \\
FedPLVM & \underline{69.64} & \textbf{85.26} & 74.14 & 56.67 & 71.428 & +3.578 \\
FPL & 68.30 & \underline{84.74} & 82.76 & \underline{70.00} & \underline{76.45} & \underline{+8.6} \\
\textbf{Ours} & \textbf{72.77} & \underline{84.74} & 75.86 & \textbf{80.00} & \textbf{78.342} & \textbf{+10.492} \\
\hline
\end{tabular}
\end{table*}

\begin{table*}[!tb]
\centering
\caption{\textbf{Performance comparison on DomainNet.} Details in \cref{sec:perf_cmp}.}
\label{table3}
\begin{tabular}{l|c|c|c|c|c|c|c|c}
\hline
Methods & Clipart & Infograph & Painting & Quickdraw & Real & Sketch & Avg. & $\Delta$ \\ \hline
FedAvg & \underline{66.79} & 27.53 & 37.80 & 36.70 & 53.93 & 40.45 & 43.867 & - \\
MOON & 53.21 & 26.35 & 35.09 & 31.80 & 51.27 & 39.02 & 39.457 & -4.41 \\
FCCL+ & 61.07 & 30.57 & 36.60 & 32.60 & 51.62 & 30.28 & 40.457 & -3.41 \\
FedProx & 62.86 & 29.05 & 37.65 & 36.80 & 56.47 & 43.09 & 44.32 & +0.453 \\
FedProto & 65.71 & 27.36 & 36.14 & 46.30 & 55.08 & 41.67 & 45.377 & +1.51 \\
FPL & 65.71 & \underline{31.08} & \underline{38.55} & 45.10 & \underline{56.81} & \underline{45.33} & 47.097 & +3.23 \\
FedPLVM & 65.71 & \textbf{32.43} & 37.05 & \underline{55.50} & 54.16 & \textbf{46.75} & \underline{48.6} & \underline{+4.733} \\
\textbf{Ours} & \textbf{68.21} & 28.55 & \textbf{39.16} & \textbf{57.80} & \textbf{63.28} & 40.65 & \textbf{49.608} & \textbf{+5.741} \\
\hline
\end{tabular}
\end{table*}

\section{Experiments}

We evaluate our algorithm on Digit-5\cite{digit5}, Office-10\cite{office10} and DomainNet\cite{domainnet}, consisting $5$, $4$ and $6$ different domains respectively. As for DomainNet, we follow the setup in FedPCL\cite{fedpcl} using a 10-class subset. In each experiment, we use one client for each domain. For Digit-5 and Office-10, each client possesses about $300$ training samples. For DomainNet, each client possesses about $400$ training samples. Clients always possess all testing samples.

We compare our algorithm with classic FL methods: FedAvg\cite{fedavg}, FedProx\cite{fedprox}, MOON\cite{moon}, FCCL+\cite{fcclplus} and FedPL methods: FedProto\cite{fedproto}, FPL\cite{fpl} and FedPLVM\cite{fedplvm}.

For all algorithms, we employ the ResNet10\cite{resnet10} as the backbone model and configure the feature vectors' dimension to $512$. We use an SGD optimizer with $lr=0.01,momentum=0.9,weight\_decay=1e-5$ for optimization. Global communication rounds are fixed at $T=50$ for Digit-5 and DomainNet and $T=100$ for Office-10, and each local training epoch consists of $E=10$ iterations.

\begin{figure}[!tb]
  \centering
  \includegraphics[width=\linewidth]{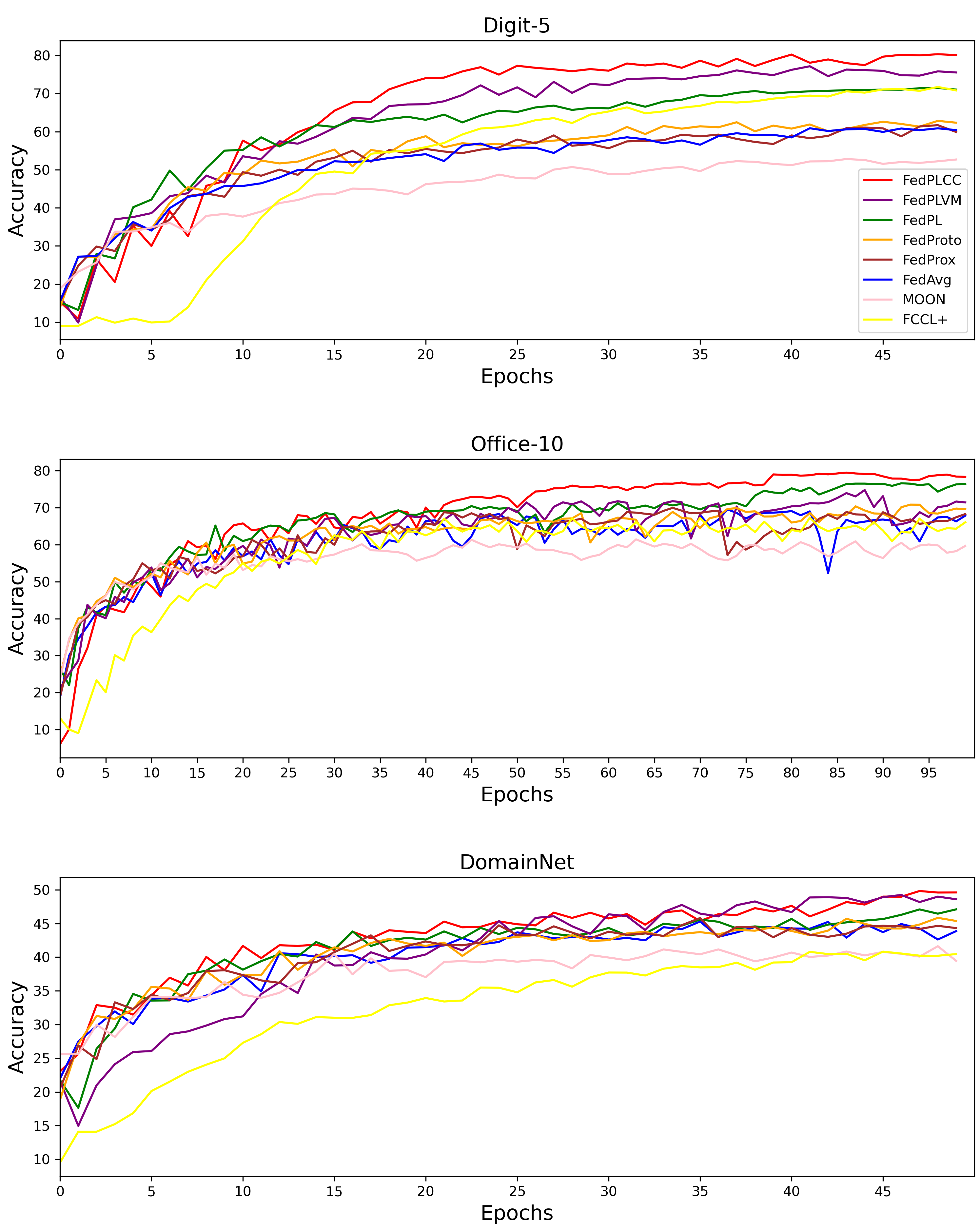}
  \caption{\textbf{Comparison of average accuracy on different communication epochs on all three experiments.} See \cref{sec:perf_cmp} for details.}
  \label{fig:6}
\end{figure}

As for hyper-parameters, we maintain $\lambda_2=10\lambda_1,\phi=0.5,\tau=0.07,\alpha=0.5$, and set $\lambda_1=100$ for Digit-5, $\lambda_1=20$ for Office-10 and $\lambda_1=1$ for DomainNet. We have done ablation experiments in \cref{sec:ablstu} to explain the chosen hyper-parameters. We have also slightly adjusted some hyper-parameters in other methods so that they could perform the best under our experiment settings. Our experiment results are presented in \cref{table1,table2,table3}, and the average accuracy metric in each communication epoch during the training phase is shown in \cref{fig:6}.

\subsection{Performance Comparison}
\label{sec:perf_cmp}

Our method demonstrates superior average accuracy across all datasets and consistently achieves higher accuracy than existing state-of-the-art methods on numerous sub-datasets. The significant accuracy improvements observed on the Digit-5 and Office-10 datasets are particularly noteworthy. For instance, on the Office-10 dataset, the variance among the four test results is merely 11.97\%, underscoring the robustness and effectiveness of our approach across datasets of varying difficulty levels.

However, our method's performance improvement on the DomainNet dataset is more modest. We attribute this to the local models' reaching their limits regarding feature extraction capabilities. Consequently, federated prototype learning algorithms that leverage contrastive feature information cannot achieve substantially better results in this scenario.

\subsection{Ablation Study}
\label{sec:ablstu}

We performed a series of ablation studies using the three datasets (mainly the Digit-5) to evaluate each component's effectiveness in our proposed method.

\subsubsection{Impact of key components. }
\label{sec:abl_key}

We conducted ablation experiments on several key components of our algorithm using the Digit-5 dataset, and the results are presented in \cref{table4}. Removing both contrastive information losses renders our method equivalent to FedAvg, which serves as the baseline for this study. 

\begin{table}[!tb]
\centering
\caption{\textbf{Ablation study of key components on Digit-5.} w/o means we remove this term from our loss function, w/ means we keep it unchanged, $w=1$ means we set the weight of all the prototypes to $1$ when calculating the loss function, and $\phi=1$ means we set $\phi$ to $1$ when calculating the loss function. The best performance is highlighted via \textbf{bold}. Details in section \cref{sec:abl_key}.}
\label{table4}
\begin{tabular}{c|c|c|c}
\hline
$\mathcal{L}_{contra}$ & $\mathcal{L}_{corr}$ & Avg. & $\Delta$ \\ \hline
 w/o & w/o & 60.394 & - \\
 w/o & w/ & 53.496 & -6.898 \\
 w/o & $\phi=1$ & 47.81 & -12.584 \\
 w/ & w/o & 72.82 & +12.426 \\
 w/ & $\phi=1$ & 78.304 & +17.91 \\
 $w=1$ & $w=1$ & 76.628 & +16.234 \\
 \textbf{w/} & \textbf{w/} & \textbf{80.092} & \textbf{+19.698} \\
\hline
\end{tabular}
\end{table}

Subsequently, we evaluated the model's performance relative to the baseline when modifying or removing each loss individually. Retaining only $\mathcal{L}_{contra}$ led to a notable 12.426\% increase in accuracy. Conversely, retaining only $\mathcal{L}_{corr}$ resulted in lower accuracy compared to FedAvg, exacerbated when incorporating the $\phi$. This outcome underscores that focusing solely on increasing intra-class similarity impedes the stable formation of prototypes, thus hindering the learning of contrastive information. Our other experiments underscored the importance of weighting and prototype selection in enhancing model accuracy.

\subsubsection{Impact of \texorpdfstring{$\alpha$}{alpha} and \texorpdfstring{$\tau$}{tau}.}
\label{sec:abl_alpha}

\begin{figure}[!bt]
  \centering
  \includegraphics[width=\linewidth]{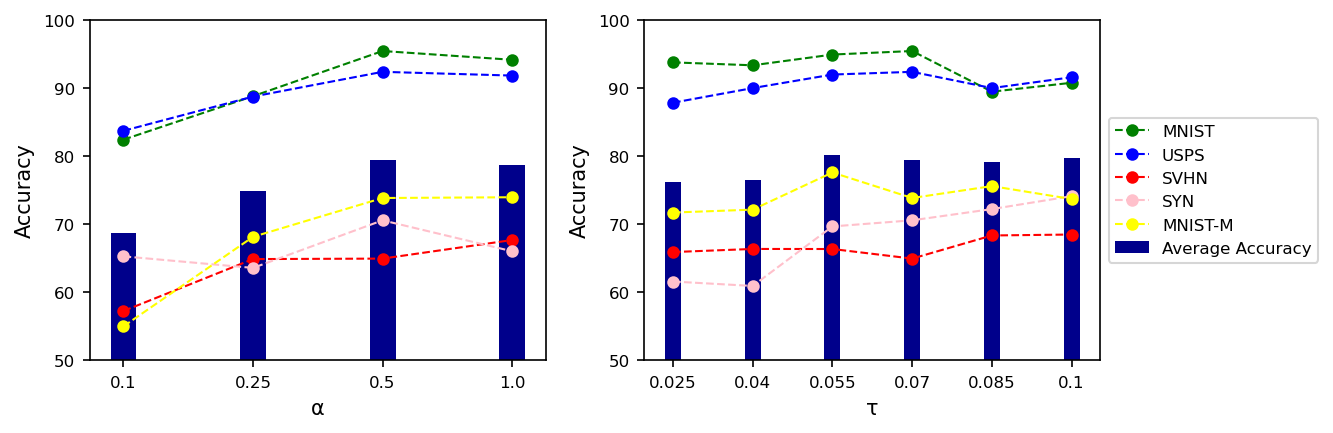}
  \caption{\textbf{Impact of $\alpha$ and $\tau$ on the Digit-5 experiment.} See \cref{sec:abl_alpha} for details.}
  \label{fig:3}
\end{figure}

FedPLVM\cite{fedplvm} includes two hyper-parameters, $\alpha$ and $\tau$, in its inter-class loss component. $\alpha$ dictates the alignment strength between local features and the global prototype, and $\tau$ determines the strength of prototype aggregation for the same label. FedPLVM discusses these two hyper-parameters and determines that $\alpha=0.25$ and $\tau=0.07$ are the optimal values. We also experiment with our method. Results are shown in \cref{fig:3}.

In our experiments, we find that setting $\alpha=0.5$ yields the highest accuracy across our datasets, and this trend is consistent across various sub-datasets. This optimal $\alpha$ value is higher than the one identified in FedPLVM, indicating that our approach does not enforce strict alignment between features and prototypes. This aligns with our objective of allowing features to naturally converge into their respective clusters without being overly constrained by prototype alignment. Similarly, our optimal $\tau$ value is $\tau=0.55$, which is lower than the optimal $\tau$ value reported in FedPLVM. This adjustment is primarily influenced by the changes in the $\alpha$ value.

\subsubsection{Impact of \texorpdfstring{$\phi$}{phi}. }

The parameter $\phi$ determines the proportion of prototypes considered relevant to the current feature when calculating the inter-class loss. Hence, as the domain gap between different clients decreases, a larger $\phi$ is preferable, and vice versa. We believe conducting ablation experiments for each dataset could yield better-performing $\phi$, but this does not align with our expectations. We fix this parameter at $0.5$ to ensure consistent performance of our method across different datasets, including those not explicitly tested in our experiments. The experiments in \cref{sec:abl_key} have already demonstrated that the top-k mechanism improves the results, which is sufficient for our purposes here.

\subsubsection{Impact of \texorpdfstring{$\lambda_2$}{lambda 2}. }
\label{sec:abl_lam2}

\begin{figure}[!tb]
  \centering
  \includegraphics[width=0.8\linewidth]{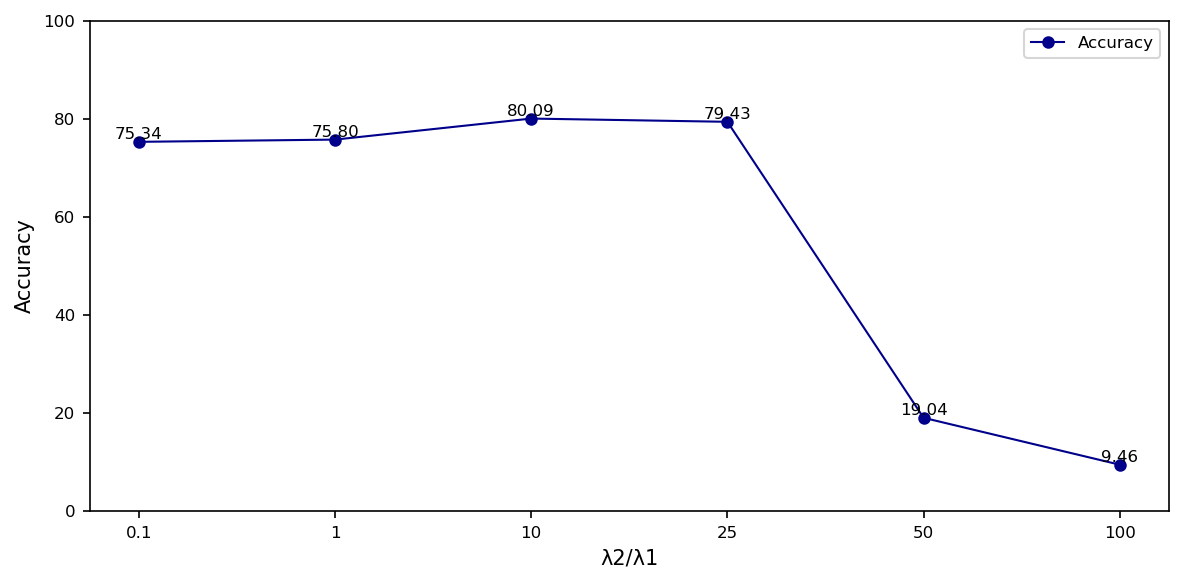}
  \caption{\textbf{Impact of $\frac{\lambda_2}{\lambda_1}$ on the Digit-5 experiment.} See \cref{sec:abl_lam2} for details.}
  \label{fig:5}
\end{figure}

$\lambda_2$ governs the relative strength between push and pull forces in our method. To ensure adaptability across different datasets, we conduct ablation experiments on the ratio  $\frac{\lambda_2}{\lambda_1}$ rather than on $\lambda_2$ alone. The experiments are specifically carried out on the Digit-5 dataset, and the results are depicted in \cref{fig:5}. Therefore, we set $\lambda_2=10\lambda_1$ in our experiment.

\subsubsection{Impact of \texorpdfstring{$\lambda_1$}{lambda 1}. }
\label{sec:abl_lam1}

\begin{figure}[!tb]
  \centering
  \includegraphics[width=\linewidth]{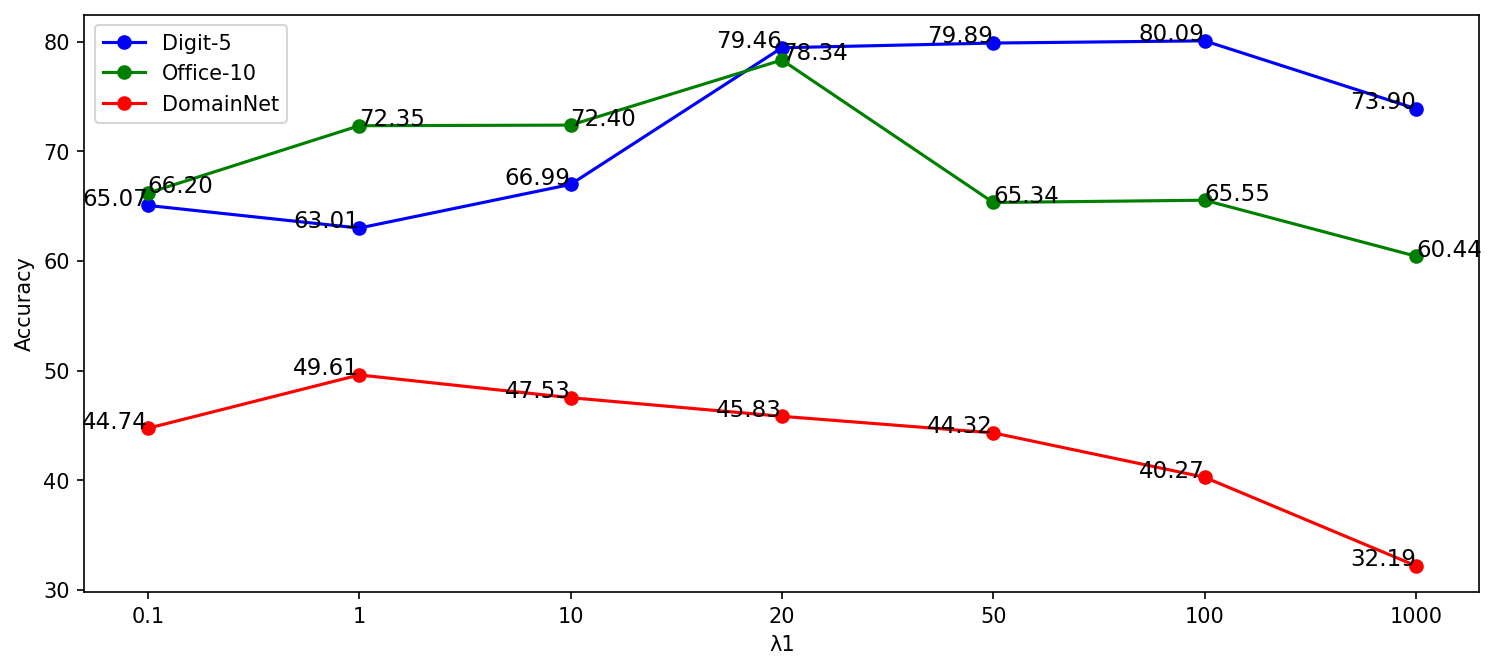}
  \caption{\textbf{Impact of $\lambda_1$ on all three experiments.} See \cref{sec:abl_lam1} for details.}
  \label{fig:4}
\end{figure}

The parameter $\lambda_1$ governs the ratio of contrastive information to label information in our method. We acknowledge that the optimal value of $\lambda_1$ can vary across different datasets. Therefore, we conducted ablation experiments across all three datasets in our study to ensure our method performs optimally and to accurately assess its effectiveness. The experimental results are depicted in \cref{fig:4}. Notably, our experiments show that setting $\lambda_1$ between 20 and 100 achieves near-optimal performance on the Digit-5 dataset. This range suggests a balanced incorporation of contrastive information and label information, highlighting the robust capability of our model in effectively extracting contrastive information.

\section{Conclusion}
In this paper, we systematically compare and summarize previous works in Federated Prototype Learning while highlighting opportunities for enhancing the clustering process. Our novel approach, FedPLCC, extends the dual-level clustering framework pioneered by FedPLVM. Through strategic weighting and selecting prototypes, as well as a redesigned loss function, our method mitigates the risk of cross-domain feature alignment, thereby facilitating a more organic convergence into distinct clusters. Experiments conducted on multiple datasets demonstrate the superiority of our approach.

\section*{Acknowledgment}

We would like to express our sincere gratitude to the Institute of Automation, Chinese Academy of Sciences, and Southern University of Science and Technology for their support and assistance in our research project.


\bibliographystyle{plain}
\bibliography{main}

\end{document}